\pdfoutput=1
\documentclass[11pt]{article}

\usepackage[final]{acl}

\usepackage{times}
\usepackage{latexsym}

\usepackage[T1]{fontenc}
\usepackage[utf8]{inputenc}

\usepackage{microtype}

\usepackage{inconsolata}

\usepackage{times}
\usepackage{latexsym}

\usepackage{microtype}

\usepackage{amssymb}
\usepackage{bm}
\usepackage{times}

\usepackage{graphicx}
\usepackage{multirow}
\usepackage{enumerate}
\usepackage{booktabs}
\usepackage{enumitem}
\usepackage{amsmath}
\usepackage{hyperref}
\usepackage{xcolor}

\author{ Jiaqi Chen$^{1}$~~~~~~~ Bingqian Lin$^{2}$~~~~~~~ Ran Xu$^{3}$ ~~~~~~~ Zhenhua Chai$^{3}$ \\ {\bf Xiaodan Liang}$^{2}$ ~~~~~  {\bf Kwan-Yee K. Wong$^{1}$}  \\
{$^1$The University of Hong Kong} \\{$^2$Shenzhen Campus of Sun Yat-sen University $^3$Meituan} \\
{\tt\small Project: \href{https://chen-judge.github.io/MapGPT/}{https://chen-judge.github.io/MapGPT/} }\\
}

\title{MapGPT: Map-Guided Prompting with Adaptive Path Planning for Vision-and-Language Navigation}

\begin{document}
\maketitle

\begin{abstract}
Embodied agents equipped with GPT as their brains have exhibited extraordinary decision-making and generalization abilities across various tasks. However, existing zero-shot agents for vision-and-language navigation (VLN) only prompt GPT-4 to select potential locations within localized environments, without constructing an effective ``global-view'' for the agent to understand the overall environment. In this work, we present a novel \textbf{map}-guided \textbf{GPT}-based agent, dubbed \textbf{MapGPT}, which introduces an online linguistic-formed map to encourage global exploration.
Specifically, we build an online map and incorporate it into the prompts that include node information and topological relationships, to help GPT understand the spatial environment. 
Benefiting from this design, we further propose an adaptive planning mechanism to assist the agent in performing multi-step path planning based on a map, systematically exploring multiple candidate nodes or sub-goals step by step. 
Extensive experiments demonstrate that our MapGPT is applicable to both GPT-4 and GPT-4V, achieving state-of-the-art zero-shot performance on R2R and REVERIE simultaneously ($\sim$10\% and $\sim$12\% improvements in SR), and showcasing the newly emergent global thinking and path planning abilities of the GPT.
% \footnote{Project: \href{https://chen-judge.github.io/MapGPT/}{https://chen-judge.github.io/MapGPT/}}

\end{abstract}

\section{Introduction}
\label{sec:intro}

\begin{figure}[t]
\begin{center}
 \includegraphics[width=1.0\linewidth]{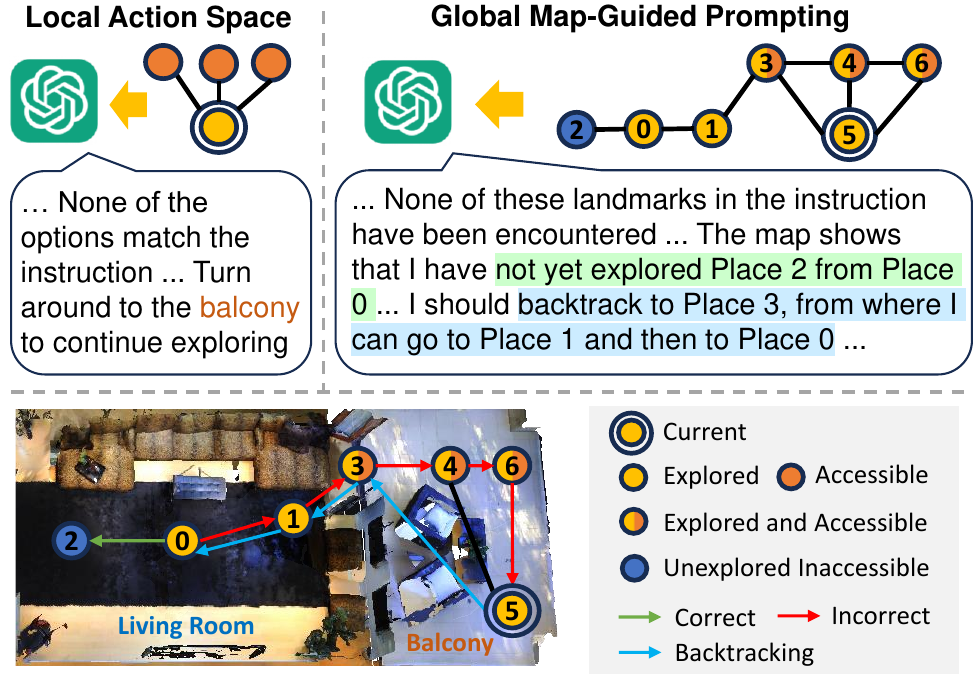}
\end{center}
\vspace{-5mm}
  \caption{
  A comparison of the thinking process of the GPT agent without and with topological maps. Given only a local action space, the agent may explore aimlessly, especially when navigation errors have already occurred. Incorporating topological maps enables the agent to understand spatial structures and engage in global exploration and path planning.
  }
\vspace{-5mm}
\label{fig:intro}
\end{figure}

% LLM background, GPT-based agents
Large language models (LLMs) \cite{touvron2023llama,touvron2023llama2, chowdhery2022palm,anil2023palm} have demonstrated strong performance in various domains.
As the most powerful LLMs, the GPT series models \cite{brown2020language,openai2023gpt4,openai2023gpt4v-system,openai2023gpt4v-technical} can even serve as the brains of embodied agents \cite{huang2023voxposer,mu2023embodiedgpt,ahn2022can}, enabling them to engage in explicit thinking and decision-making process.
Moreover, these GPT-based agents are typically zero-shot or few-shot, eliminating the burdensome tasks of data annotation and model training, and they also demonstrate remarkable sim-to-real abilities.

% VLN task
Recently, LLMs have also been adopted in vision-and-language navigation (VLN), where agents are given human instructions that require them to visually perceive and navigate an indoor environment.
Previous learning-based VLN methods \cite{anderson2018vision,fried2018speaker,qi2020Object,chen2022think,qiao2022hop,wang2023scaling} relied on the training on large-scale domain-specific data with expert instruction annotations to execute navigation tasks.
To address their reliance on training data and potential sim-to-real gap, some GPT-based zero-shot agents \cite{zhou2023navgpt,long2023discuss} with explainable decision-making abilities have been proposed. These methods translated visual observations into textual prompts and required GPT-4 to act as an agent to select the correct position or direction. Besides, they also employed multiple additional experts to handle various subtasks in text form, such as summarizing the history and fusing repeated expert predictions.

However, such zero-shot VLN agents face some challenges. 
Firstly, these methods are developed for a multi-expert pure language system based on GPT-4, necessitating the conversion of visual observations into text and multiple rounds of text summarizations, which inevitably lead to information loss and hinder the application to multimodal LLMs (e.g., GPT-4V). 
More importantly, these agents make decisions based solely on the observations of the local environment.
As shown in Figure~\ref{fig:intro} (left), given only the local action space, when an agent realizes that it has engaged in an erroneous exploration, it can only continue to explore surrounding environment aimlessly.

In this paper, we propose MapGPT which contains a topological map in the linguistic form to assist in global exploration and adaptive path planning. 
We first develop a simple yet efficient prompt system with only one navigation expert that can be applied to both GPT-4 and multimodal GPT-4V flexibly.
To encourage global exploration, we propose a map-guided prompting method for the GPT model, to build a ``global-view'' for the agent. 
Specifically, for an online constructed map, we have discarded the precise GPS coordinates that are difficult for GPT to understand, while preserving the topological relationships between nodes and incorporating them into prompts to assist in understanding the navigation environment,
as shown in Figure~\ref{fig:intro} (right).
Given this tailored map, we further propose an adaptive planning mechanism to activate GPT's multi-step path planning ability. 
Instead of documenting the thinking process of each step as in previous works, MapGPT generates a multi-step planning similar to a human work plan, and updates it iteratively to achieve strategic exploration of potential objectives.
Benefiting from this, the agent can adaptively perform path planning based on the map, systematically explore multiple candidate nodes or sub-goals step by step, and backtrack to a specific node for re-exploration when necessary.

We conduct experiments on two popular VLN benchmarks, namely R2R~\cite{anderson2018vision} and REVERIE~\cite{qi2020reverie}, which contain step-by-step and high-level instructions respectively. Experimental results show the superiority of MapGPT over existing zero-shot VLN agents. Especially in REVERIE, MapGPT exhibits enhanced competitiveness (31.6\% SR), surpassing even some learning-based methods trained on REVERIE. Extensive ablation studies reveal the advantage of our introduced map and adaptive path planning mechanism in encouraging systematic exploration to improve navigation. 

Our contributions can be summarized as follows.
\begin{itemize}
\setlength{\itemsep}{0pt}
\setlength{\parsep}{0pt}
\setlength{\parskip}{0pt}
  \item We propose a novel map-guided prompting method, which introduces an online linguistic-formed map including node information and topological relationships to encourage GPT's global exploration.
  \item
  An adaptive planning mechanism is utilized to activate GPT's multi-step path-planning ability, enabling systematic exploration of multiple potential objectives.
  \item MapGPT can be applied to both GPT-4 and GPT-4V and is more unified as it can adapt to varying instruction styles effortlessly, achieving state-of-the-art zero-shot performance on both the R2R and REVERIE datasets.

\end{itemize}
\section{Related Work}

\noindent\textbf{Vision-and-Language Navigation (VLN)}
As a representative multi-modal embodied AI task, VLN requires an agent to combine human instructions and visual observations to navigate and locate targets in real-world scenes.
Previous learning-based approaches~\cite{wang2019reinforced,ma2019self, deng2020evolving,qi2020Object} proposed various model architectures and trained their models on domain-specific datasets. Besides, pretrained models~\cite{hong2021vln,chen2021history,chen2022think,qiao2022hop,guhur2021airbert,an2022bevbert,lin2022adapt,qiao2023march,wang2023scaling,pan2023langnav} have been widely applied to produce better multi-modal representations.
Recently, to address the reliance on domain-specific data and the possible sim-to-real gap, some zero-shot agents based on GPT~\cite{zhou2023navgpt,long2023discuss} have been proposed. However, they suffer from several limitations.
For example, NavGPT \cite{zhou2023navgpt} has limited performance and relies on a two-stage language-only system. 
DiscussNav \cite{long2023discuss} introduces a sequential multi-experts system to discuss and summarize various information and fuses five repeated predictions to improve performance.
Some of their designs limit the agent's capability to only address step-by-step instructions in the R2R dataset \cite{anderson2018vision}, and have not been validated on other styles of instructions (e.g., REVERIE \cite{qi2020reverie}). Besides, these agents are limited to local exploration as they can only reason and make decisions within adjacent navigable points.
In this paper, we propose a map-guided prompting method with adaptive path planning for global exploration, achieving impressive performance on both R2R and REVERIE.

\noindent\textbf{Large Language Models (LLMs)}
LLMs \cite{openai2023gpt4,openai2023gpt4v-system,openai2023gpt4v-technical,vicuna2023,touvron2023llama,touvron2023llama2,anil2023palm} have demonstrated remarkable capabilities in multiple domains. Recently, LLM-based agents \cite{huang2023voxposer,mu2023embodiedgpt,ahn2022can,pan2023langnav,brohan2023rt,schumann2023velma,hu2023look,lin2024navcot} have also attracted significant interest of the community.
For example, 
VoxPoser \cite{huang2023voxposer} utilizes LLM and vision-language models to extract affordances and constraints, which enables motion planners to generate trajectories for manipulation.
LangNav \cite{pan2023langnav} employs LLMs for navigation, but it merely utilizes GPT-4 \cite{openai2023gpt4} to synthesize some data and performs fine-tuning using Llama2 \cite{touvron2023llama2} as the backbone, rather than directly employing LLM as a zero-shot agent.
In fact, the application of LLM-based zero-shot agents in navigation tasks is still limited, and how to prompt LLMs (including multimodal GPT-4V) to activate global thinking and planning abilities required by navigation task have yet to be explored~\cite{yang2023dawn}.

\noindent\textbf{Maps for Navigation}
Maps used for navigation tasks can be primarily categorized into two types, i.e., metric maps and topological maps.
Employing SLAM~\cite{fuentes2015visual} for constructing metric maps~\cite{chaplot2020object,thrun1998learning} is widely used in navigation. 
However, this type of approach requires a trade-off between map size and computational efficiency, which affects navigation performance.
To address this limitation, graph-based topological maps \cite{chen2021topological,chen2022think,an2022bevbert} have been proposed for pre-exploring environment or enabling global exploration, such as backtracking to previously visited nodes. 
However, these methods are all designed for model learning. It remains unexplored how to build a map with prompts and leverage the powerful capabilities of LLMs for zero-shot reasoning and planning based on the map.

\begin{figure*}[t]
\begin{center}
 \includegraphics[width=0.96\linewidth]{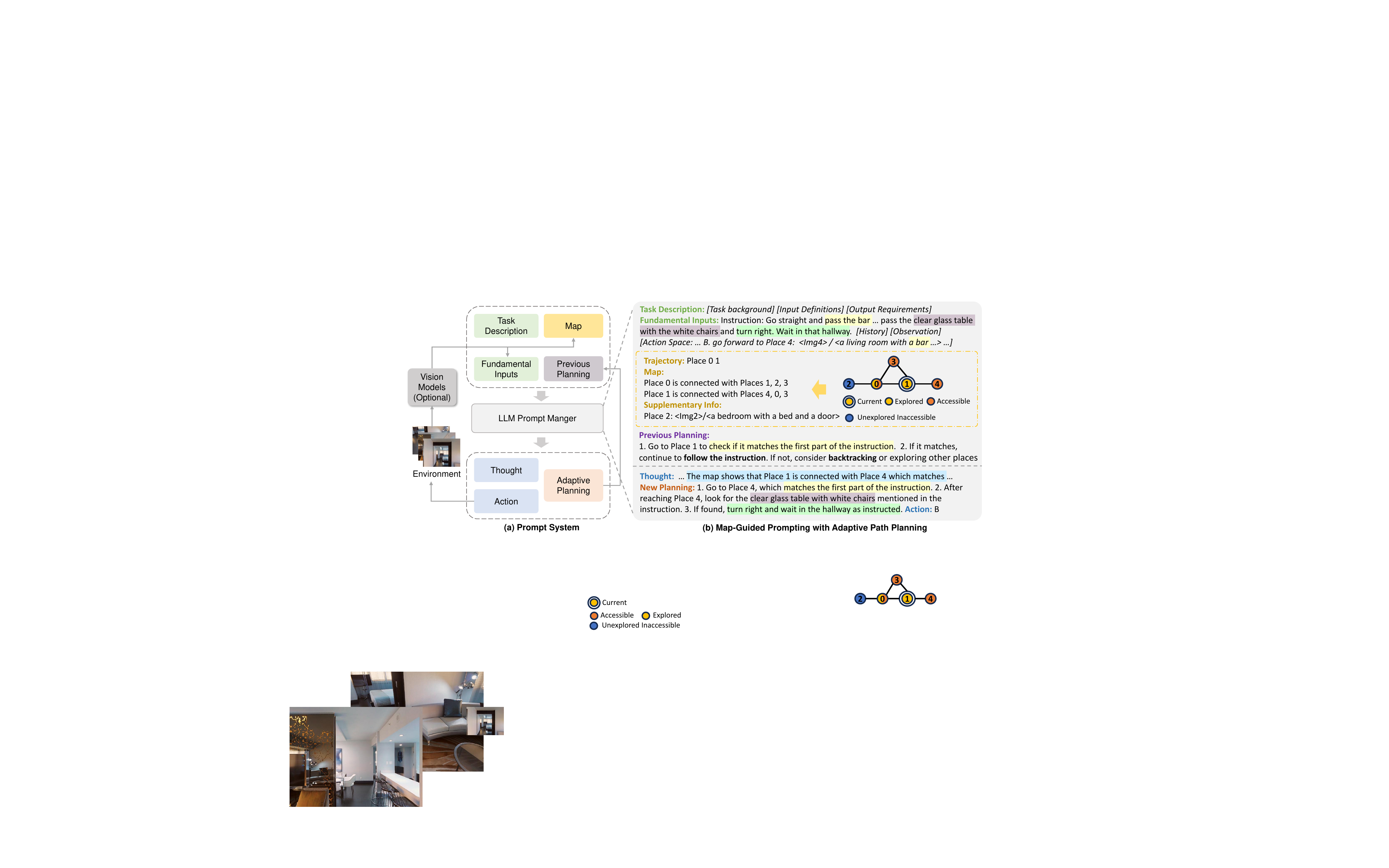}
\end{center}
\vspace{-6mm}
  \caption{
  % The framework of our proposed MapGPT. 
  Our basic system consists of two types of prompts, namely task description and fundamental inputs.
  We introduce a map-guided prompting method that builds an online-constructed topological map into prompts, activating the agent's global exploration. We further propose an adaptive mechanism to perform multi-step path planning on this map, systematically exploring candidate nodes or sub-goals. Note that vision models are optional, and viewpoint information can be represented using either the image or textual description of the observations.
  }
\vspace{-5mm}
\label{fig:framework}
\end{figure*}

\section{Method}

In this section, we introduce our newly designed prompt system (Sec.~\ref{sec:system}), the details of our map-guided prompting method to help the agent understand global environment (Sec.~\ref{sec:map}), and the novel adaptive mechanism that encourages the agent to make multi-step path planning (Sec.~\ref{sec:planning}).

\subsection{Single Expert Prompt System}
\label{sec:system}
Previous works such as NavGPT \cite{zhou2023navgpt} and DiscussNav \cite{long2023discuss} are two-stage systems. They first gather visual observations from all the views and translate them into textual descriptions which are then fed into language-only GPT-4~\cite{openai2023gpt4} for decision-making. Besides, they rely on complex multi-expert designs, where GPT plays different roles to achieve various functions, such as instruction parsing, summarizing textual descriptions and history, etc. 
However, these intricate designs are only geared towards text processing, which limits their research value. Powerful multimodal large models, such as GPT-4V~\cite{openai2023gpt4v-system,openai2023gpt4v-technical}, can directly serve as the agent's brain to process multimodal information and make decisions~\cite{yang2023dawn}.

Compared with previous works, our proposed single expert prompt system has several unique features. 
(1) We eliminate the need for a separate design of an additional historical summary expert or instruction decomposing expert based on GPT models, which makes it convenient to incorporate both visual/textual inputs and additional information, such as maps.
(2) Our navigation expert can utilize GPT-4V to make decisions directly based on visual observations in one stage. It can also take text descriptions as inputs and flexibly apply them in the two-stage GPT-4 system.
(3) Simple yet efficient. In the R2R dataset, our two-stage system requires an average of 672 input tokens and 115 output tokens per step. In comparison, NavGPT utilizes three GPT experts, resulting in an average cost of 2,465 input tokens and 317 output tokens per step.

As shown in Figure~\ref{fig:framework}, we collect various fundamental inputs for the agent, including instruction $I$, history $H_t$, observation $O_t$, and action space $A_t$. The meaning of these inputs, as well as the requirements for output, are clearly pre-defined in the task description $D$. We utilize a prompt manager $PM$ to organize these prompt inputs which are then fed into the large language model $LLM$ to generate the current thought $T_t$ and select a specific action $a_{t,i} \in A_t$. The pipeline can be formulated as
\begin{equation}
    T_t, a_{t,i} = LLM(PM(D, I, H_t, O_t, A_t)).
\end{equation}
% Details will be provided in the following subsections.

\subsubsection{Task Description}
As shown in Figure \ref{fig:framework}, our prompts for task description $D$ consist of three main parts. We set the task background for GPT, provide some definitions regarding the inputs at each step, and propose some basic requirements regarding how the agent should accomplish this task.

\subsubsection{Fundamental Inputs}

\paragraph{Instruction $I$}
Since we aim to adapt both fine-grained and high-level instructions, we directly feed the raw instructions to GPT without any initial analysis or decomposition.

\paragraph{Visual Observation $O_t$}
Our agent is equipped with an RGB camera to capture $M$ images of the environment. 
Unlike NavGPT~\cite{zhou2023navgpt} that gathers excessive environmental information, we prioritize the observations of navigable points, which can be formulated as $O_t=\{o_{t,i}\}_{i=1}^N$, where $N$ is the number of navigable viewpoints at step $t$. Each $o_{t,i}$  represents the observation towards a specific navigable point. 
These observations are original images for a one-stage system based on GPT-4V and can also be replaced with caption and detection results in a two-stage system, where we follow NavGPT \cite{zhou2023navgpt} and utilize off-the-shelf vision models BLIP-2 \cite{li2023blip} and Faster R-CNN \cite{ren2015faster,anderson2018bottom} to acquire scene description and object detection respectively. 
NavGPT also utilizes the bounding boxes provided by the REVERIE dataset to extract an additional object list of all the views, which is then used for the R2R experiment. We also utilize this object list and name it the ``Surroundings''.
However, we do not utilize this additional information in R2R. We only utilize it to enable the agent to determine whether to stop in the REVERIE dataset since we do not specify the directions of these surrounding objects.

\paragraph{Action Space $A_t$}
NavGPT allows the agent to directly select a viewpoint and DiscussNav enables the model to predict a direction. In this paper, we incorporate both directional phrases  (e.g., turn left to) and the corresponding observations of each direction into the action space for GPT to select from, making the decision-making process more intuitive. 

Specifically, we take as an input the action space $A_t=\{a_{t,i}\}_{i=0}^{N}$ for GPT model.
$N$ is also the number of navigable points and we additionally define $a_{t,0}$ as \textit{``A. stop''} so that the agent has $N+1$ options in total. Each $a_{t,i}$ in the remaining $N$ options is formulated using the template:
\begin{quote}
\centering
\textit{``\{label\} \{direction\} \{$o_{t,i}$ \}''}.    
\end{quote}
At each step $t$, the agent only needs to choose one option $a_{t,i} \in A_t$. For the output format, we require the agent to  simply provide a single option label, such as \textit{"Action: B"}.

\paragraph{History $H_t$}
We record all previous actions $a_0 \sim a_{t-1}$ for history. The following prompt template is utilized for appending the actions into $H_t$:

\begin{quote}
\centering
\textit{``step 0: \{$a_0^\ast$\}, ..., step t-1: \{$a_{t-1}^\ast$\}''}, 
\end{quote}
in which $t \geq 1$ and $a^\ast$ denotes the selected action $a$ but with the option label removed. 
The initial history is defined as $H_0$ = \textit{``The navigation has just begun, with no history''}.

\subsection{Map-Guided Prompting}
\label{sec:map}

For the VLN task, previous work \cite{zhu2021soon,chen2022think,an2023bevbert} has demonstrated the effectiveness of online constructed maps for global navigation. However, how to construct the maps and transform them into a certain form to prompt LLMs has not been investigated in this domain. 
Additionally, we observe that GPT-4V struggles slightly in its attempts to understand navigation environments based on multiple precise coordinates.
Therefore, we propose a novel map-guided prompting approach that converts topological relationships of a map into textual prompts, as shown in Figure~\ref{fig:framework}(b).

\paragraph{Topological Mapping}
In the VLN task, the agent has never explored the entire environment and must construct a map based on its own observations online. We store the map as a dynamically updated graph, following the graph-based method DUET \cite{chen2022think}. At each step $t$, we record all observed nodes along the navigation trajectory and their connectivity into the graph $G_t = \{V_t, E_t\}$, where $V_t=\{v_{t,i}\}_{i=1}^{K}$ are a series of $K$ observed nodes marked with the index $i$ in the order of observations. All the edges between these observed nodes are recorded in $E_t$.
At any step $t$, given the current location, the simulator will provide several neighboring nodes that are currently navigable. These new nodes and edges will be utilized for updating the graph from $G_{t-1}$ to $G_t$.

\subsubsection{Constructing Maps with Prompts}

After we have obtained a topological connectivity graph representing the structure of the environment, the next step is to transform it into an appropriate prompt and add map annotations to form a complete map-guided prompt to help the agent understand the navigation environment.
For each step $t$, we categorize all observed nodes in the environment into three types, namely (1) explored nodes $\{en_{j}\}_{j=0}^{t}$ (including starting node $en_0$ and current node $en_t$), (2) accessible nodes $\{an_t^0, an_t^1, ... \}$, and (3) unexplored inaccessible nodes $\{un_0, un_1, ... \}$. 

\paragraph{Trajectory}
As we have already marked each location during the navigation process, it follows logically that we can create a simplified trajectory prompt to help the agent  understand its navigation path in the map and avoid repeated exploration as far as possible. For the explored nodes, we formulate our trajectory prompt using the template:
\begin{quote}
\centering
    \textit{``Trajectory: Place \{$en_0$\} ... \{$en_t$\}''},
\end{quote}
where each $en_{j}$ corresponds to a node $v_{t,i} \in V_t$ stored in the order of observation. Thus, we consider the index $i$ as the ID of the node and fill it into the template to denote $en_j$.

\paragraph{Map Connectivity}
Unlike DUET, which also converts precise GPS coordinates into embeddings for graph learning, we only retain the topological relationships of the map nodes since we discover that it seems challenging for GPT to understand precise coordinate data. These topological relationships are transformed into textual prompts, making it easier to comprehend spatial structures.
Since the connectivity can only be observed at explored nodes, we always start with \textit{``Place \{$en_t$\} is connected with ...''}. All IDs corresponding to the neighboring accessible nodes of these exploded nodes will be listed using the following template:

\vspace{3mm}
    \noindent \textit{``Map:} \\
    \textit{Place \{$en_0$\} is connected with Places \{$an_0^0$\}, ...} \\
    \textit{Place \{$en_1$\} is connected with Places \{$an_1^0$\}, ...} \\
    ... \\
    \textit{Place \{$en_t$\} is connected with Places \{$an_t^0$\}, ...''},
\vspace{3mm}

\noindent where all the nodes should be filled with their node IDs. Note that this map connectivity does not need to be updated if the agent decides to backtrack and revisit some previously explored nodes.

\paragraph{Map Annotations}
The final step involves adding an annotation to each node of this topological map, enabling the agent to refer to them for path planning. As we have already provided currently accessible nodes in the action space, and the selected actions are also included in the history to form explored nodes, there is no need to repeat them. It is sufficient to simply add the node IDs in the action space at each step. Specifically, each $a_{t,i}$ in the action space $A_t$ is reformulated as

\vspace{3mm}
\textit{``\{label\} \{direction\} Place \{$an_t^i$\}: \{$o_{t,i}$ \}''}.
\vspace{3mm}

The agent can therefore find the corresponding explored nodes and accessible nodes in history $H_t$ and action space $A_t$ respectively. However, we still have some unexplored and currently inaccessible nodes that are important, especially when the agent encounters obstacles in exploration and needs to revisit previous nodes for re-exploration. 
These inaccessible nodes are considered as supplementary information to assist the agent in backtracking to the most suitable node. In \textit{``Supplementary Info''} prompts, we record their raw images for the one-stage system, whereas in the two-stage system, it can be replaced with the corresponding scene descriptions.

\subsection{Adaptive Path Planning}
\label{sec:planning}

NavGPT and DiscussNav record the thinking process of the agent at each step. Despite employing another GPT expert for summarization, they still involve a significant amount of redundancy. This is also not consistent with human thinking, as we usually do not document every moment of our thoughts. Instead, we tend to document a work plan and update it as necessary. 

Inspired by the above insight and benefiting from the utilization of maps, we propose an adaptive planning module that demands the agent to dynamically generate and update its multi-step path planning at each step.
Concretely, the agent is required to combine the thought, map, and previous planning to adaptively update a new multi-step path planning. The entire process is iterative, where the planning output of the current step serves as the input to the next step, allowing the agent to refer to previous plans.
Therefore, at step $t$, the agent should refer to the last planning $P_{t-1}$, 
where $P_0$ is set as \textit{``Navigation has just started, with no planning yet''}.

In summary, the proposed MapGPT that combines map $M_t$ and path planning $P_{t-1}$ can be defined as follows:
\begin{equation}
% \small
\begin{split}
    T_t, P_t, a_{t} = & LLM(PM(D, I, H_t, O_t, A_t, \\
    & M_t, P_{t-1})).
\end{split}
\end{equation}
In addition to the widely-used thought and action on the previous agents, MapGPT outputs additional multi-step planning information. 
Thanks to the powerful reasoning abilities of LLMs, the agent can focus on multiple potential nodes or sub-goals during planning, instead of being limited to predicting only one optimal choice from the global action space in a probabilistic manner without interpretability, as in the previous DUET model.
Furthermore, the agent can adaptively update its plan, choosing to continue exploring sub-goals or backtrack to a previous node for re-exploration, which enhances the agent's navigation performance.
Some analysis of these capabilities is presented in Section \ref{sec:experiments}.

\begin{table}[t] 
 \centering
 \vspace{-2mm}
 \small
 \renewcommand\tabcolsep{1.0pt} % column space
\resizebox{1.0\columnwidth}{!}{
 \begin{tabular}{lcc|cccc}
 \toprule
 Methods & LLMs & Exp & NE$\downarrow$ & OSR$\uparrow$ & SR$\uparrow$ & SPL$\uparrow$  \\
 \midrule
 NavGPT (\citeauthor{zhou2023navgpt}) & GPT-3.5 & 3 & 8.02 & 26.4 & 16.7 & 13.0 \\
 MapGPT (Ours) & GPT-3.5 & \textbf{1} & 8.48 & 29.6 & 19.4 & 11.6   \\
 DiscussNav (\citeauthor{long2023discuss}) & GPT-4 & 5 & 6.30 & 51.0 & 37.5 & 33.3 \\
 MapGPT (Ours) & GPT-4 & \textbf{1} & 5.80 & \textbf{61.6} & 41.2 & 25.4 \\
 MapGPT (Ours) & GPT-4V & \textbf{1} & \textbf{5.62} & 57.9 & \textbf{47.7} & \textbf{38.1} \\
 \bottomrule
 \end{tabular}
 }
 \vspace{-2mm}
\caption{Results on 72 various scenes of the R2R dataset. ``Exp'' refers to the number of GPT experts.  
}
 \vspace{-2mm}
\label{table:72_scenes}
\end{table}
%%%%%%%%%%%%%%% 72 scenes %%%%%%%%%%%%%%%%%%

\section{Experiments}
\label{sec:experiments}

\subsection{Experimental Settings}

\paragraph{Datasets and Evaluation}
We choose two datasets, R2R~\cite{anderson2018vision} and REVERIE~\cite{qi2020reverie}, to validate our MapGPT since they have distinct instruction styles. R2R provides detailed step-by-step instructions while REVERIE only offers a high-level description of finding the target object, which usually requires more exploration in the environment.
To unify the prompt system, we focus only on navigation performance, which involves finding the correct location or object to stop, while neglecting the object grounding sub-task in REVERIE. We therefore adopt several evaluation metrics for navigation, including Navigation Error (NE, the distance between agent's final location and the target location), Success Rate (SR), Oracle Success Rate (OSR, SR given Oracle stop policy), and SR penalized by Path Length (SPL).

\subsection{Experimental Results}

\subsubsection{Results on the Room-to-Room Dataset}

\noindent\textbf{Various scenes.}
As shown in Table~\ref{table:72_scenes}, 
we employ an identical sampled subset (72 scenarios, 216 trajectories) as in NavGPT's experiment to evaluate the zero-shot performance across various scenarios.
In addition, some of DiscussNav's experiments fuse five repeated predictions to enhance performance. 
For a fair comparison, we evaluate the performance of our MapGPT, and previous NavGPT and DiscussNav under the single-prediction setting.
For a two-stage system, MapGPT outperforms previous models when applied to different LLMs (including GPT-3.5 and GPT-4). 
When utilizing GPT-4V as a one-stage agent, combined with the proposed map-guided prompting with adaptive path planning, MapGPT achieves a success rate of 47.7\%.
MapGPT (GPT-4 based) has limited performance on the SPL metric, which could be attributed to the fact that map-guided prompting encourages the agent to continue global exploration when encountering insufficient textual descriptions. The agent has a longer navigation path, thus impacting the SPL metric. On the other hand, MapGPT based on GPT-4V avoids the issue of information loss during vision-to-text conversion and achieves an SPL of 38.1\%.

%%%%%%%%%%%% val unseen %%%%%%%%%%%%
\begin{table}[t] 
 \centering
 \vspace{-2mm}
 \small
 \renewcommand\tabcolsep{1.0pt} % column space
\resizebox{1.0\columnwidth}{!}{
 \begin{tabular}{llcccc}
 \toprule
 Settings & Methods & NE$\downarrow$ & OSR$\uparrow$ & SR$\uparrow$ & SPL$\uparrow$  \\
 \midrule
 \multirow{3}{*}{Train} 
& Seq2Seq~\cite{anderson2018vision} & 7.81 & 28 & 21 & - \\
 & Speaker~\cite{fried2018speaker} & 6.62 & 45 & 35 & - \\
 % & Speaker-Follower~\cite{fried2018speaker} & 6.62 & 45 & 35 & - \\
 & EnvDrop~\cite{tan2019learning} & 5.22 & - & 52 & 48 \\
 \midrule
 \multirow{6}{*}{Pretrain}
 & LangNav \cite{pan2023langnav} & - & - & 43 & - \\
 & PREVALENT \cite{hao2020towards} & 4.71 & - & 58 & 53 \\
 & RecBERT \cite{hong2021vln} & 3.93 & 69 & 63 & 57  \\
  & HAMT \cite{chen2021history} & 2.29 & 73 & 66 & 61  \\
  & DUET \cite{chen2022think} &3.31 & 81 & 72 & 60 \\
 & ScaleVLN \cite{wang2023scaling} & 2.09 & 88 & 81 & 70 \\
 \midrule
 \multirow{3}{*}{ZS} 
 % & DUET (Init. LXMERT \cite{tan2019lxmert}) & 9.74 & 7 & 1 & 0 \\
 & NavGPT \cite{zhou2023navgpt} & 6.46 & 42 & 34 & 29 \\
 & MapGPT (with GPT-4) & 6.29 & 57.6 & 38.8 & 25.8 \\
 & MapGPT (with GPT-4V)  & \textbf{5.63} & \textbf{57.6} & \textbf{43.7} & \textbf{34.8} \\
 \bottomrule
 \end{tabular}
}
% \vspace{-3mm}
\caption{Results on the validation unseen set of the R2R dataset. MapGPT outperforms two non-pretrained methods and the zero-shot NavGPT.}
% \vspace{-6mm}
\label{table:r2r_unseen}
\end{table}
%%%%%%%%%%%%%%%%%%%%%%%%%%%%%%%

\noindent\textbf{Validation unseen set.}
We further compare the navigation performance between the proposed MapGPT and previous NavGPT on a larger validation unseen set with 11 scenes and 783 trajectories. As shown in Table~\ref{table:r2r_unseen}, due to the distribution difference, the success rate of MapGPT is slightly lower (38.8\% with GPT-4 and 43.7\% with GPT-4V) compared to the results on the 72 scenes. 
Compared to NavGPT (GPT-4 based), MapGPT (GPT-4 based) exhibits a noticeable reduction in NE, leading by 4.8\% in SR, and a substantial 15.6\% improvement in OSR. 
Obviously, our proposed map-guided prompting with adaptive planning has raised the upper limit of agent navigation ability (OSR), thus increasing final success rate. However, due to information loss in the two-stage pipeline, MapGPT (GPT-4 based) tends to continue exploring when encountering insufficient description and has a weaker SPL performance. Nevertheless, the GPT-4V-based agent does not suffer from this issue, thus achieving 34.8\% in SPL.

\subsubsection{Results on the REVERIE Dataset}

Benefiting from the flexible single-expert system and the utilization of a universal map-guided prompting and planning, we effortlessly apply MapGPT to the REVERIE dataset with different instruction styles.
Considering the API costs and easier comparison for future work, we randomly sample 500 instructions from the validation unseen set for the zero-shot setting REVERIE benchmark.

\noindent\textbf{Validation unseen set.}
As shown in Table~\ref{table:reverie}, MapGPT exhibits greater competitiveness on REVERIE (31.6\% SR), significantly outperforming zero-shot NavGPT and some training-only models across all metrics. Moreover, when compared to HAMT which benefits from the pretraining and fine-tuning process, MapGPT demonstrates a highly competitive performance as well. 

\noindent\textbf{Backtracking ratio.}
We analyze the trajectory of each case, and if there are repeated visits to previously visited locations, we consider it as backtracking and calculate the ratio of such occurrences in both NavGPT and MapGPT for comparison on the REVERIE dataset.
MapGPT experiences at least one instance of backtracking in 49\% of cases, and among them, there is an 80\% probability of successfully correcting its erroneous navigation path at least once. As a comparison, NavGPT, which does not utilize any maps and planning, performs poorly on this metric. Only 18\% of cases involve backtracking to visited nodes, and among these cases, 53\% of instances successfully correct the previous exploration error at least once.

%%%%%%%%%%%% REVERIE %%%%%%%%%%%%
\begin{table}[t] 
 \centering
 \vspace{-2mm}
 \small
 \renewcommand\tabcolsep{1.0pt} % column space 
 \resizebox{1.0\columnwidth}{!}{
 \begin{tabular}{llccc}
 \toprule
 Settings & Methods & OSR$\uparrow$ & SR$\uparrow$ & SPL$\uparrow$  \\
 \midrule
 \multirow{4}{*}{Train}
  & Seq2Seq~\cite{anderson2018vision}  & 8.07 & 4.20 & 2.84 \\
  & RCM~\cite{wang2019reinforced}  & 14.2 & 9.29 & 6.97 \\
  & SMNA~\cite{ma2019self} & 11.3 & 8.15 & 6.44 \\
  & FAST-MATTN~\cite{qi2020reverie}  & 28.2 & 14.4 & 7.19 \\
  \midrule
  \multirow{3}{*}{Pretrain}
  & HAMT \cite{chen2021history}  & 35.4 & 31.6 & 29.6  \\
  & DUET \cite{chen2022think}  & 50.0 & 45.8 & 35.3 \\
  & LAD \cite{li2023layout} & 64.0 & 57.0 & 37.9 \\
 \midrule
 \multirow{3}{*}{ZS}
 & NavGPT~\cite{zhou2023navgpt} & 28.3 & 19.2 & 14.6\\
 & MapGPT (with GPT-4)  & \textbf{42.6} & 28.4 & 14.5 \\
 & MapGPT (with GPT-4V)  & 36.8 & \textbf{31.6} & \textbf{20.3} \\
 \bottomrule
 \end{tabular}
 }
% \vspace{-3mm}
\caption{Comparison on a randomly sampled subset from the validation unseen set of the REVERIE dataset. Note that we have retested the released HAMT and DUET on this same subset.}
 \vspace{-1mm}
\label{table:reverie}
\end{table}

%%%%%%%%%%%% case %%%%%%%%%%%%
\begin{figure*}[t]
\begin{center}
 \includegraphics[width=1.0\linewidth]{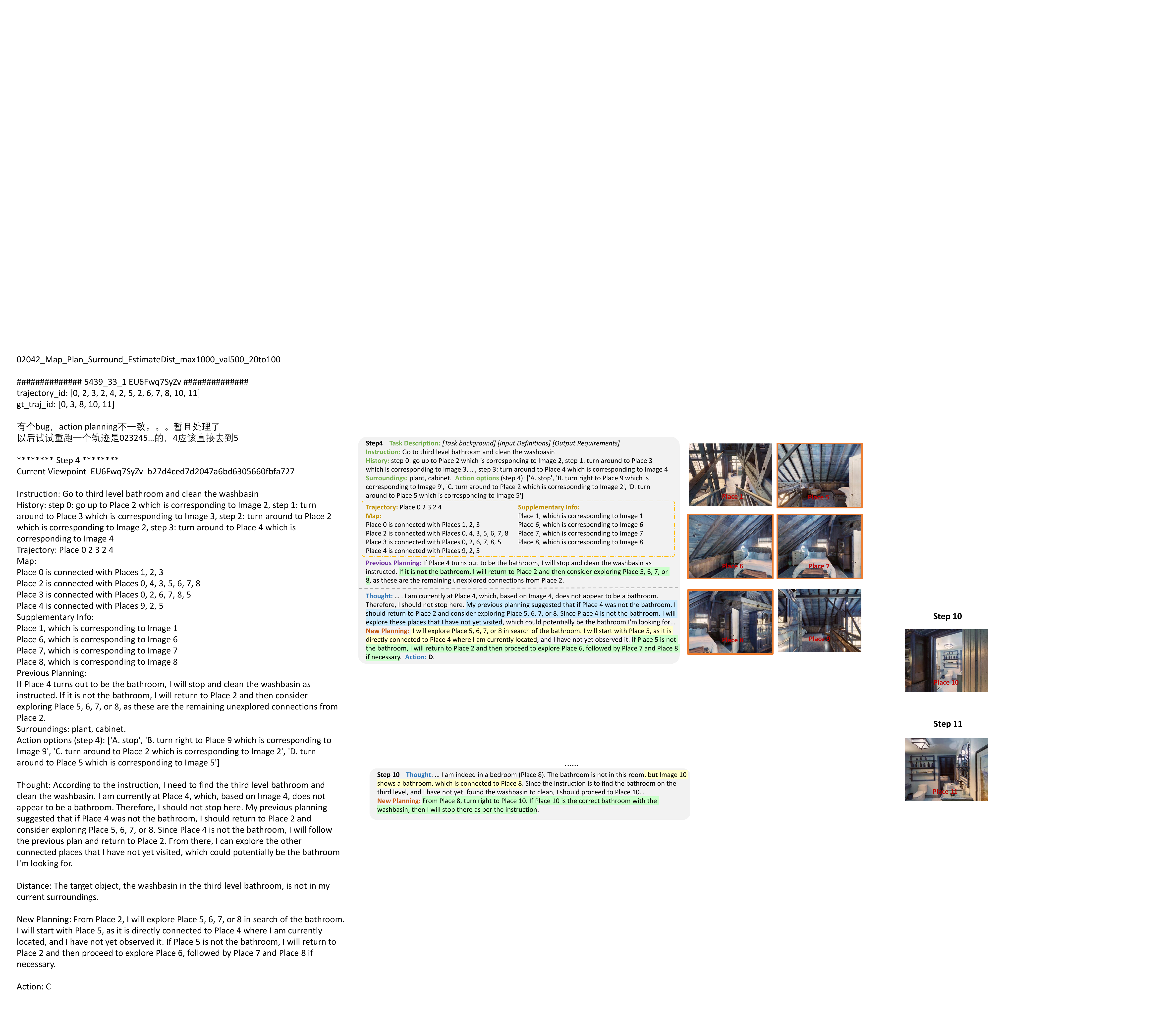}
\end{center}
\vspace{-2mm}
  \caption{
  A successful case on REVERIE showcases MapGPT's (GPT-4V based) various abilities, including global exploration (blue), map understanding (yellow), and adaptive multi-step path planning (green). The six images on the right represent six unexplored places at step 4. Among these, MapGPT focuses on four possible places and systematically explores them until it discovers the bathroom when moving to place 8.
  }
% \vspace{-5mm}
\label{fig:case}
\end{figure*}

\subsection{Ablation Study}

As shown in Table~\ref{table:ablation}, we explore the effectiveness of different prompt designs on 72 various scenes of the R2R dataset.

\noindent\textbf{Coordinate Maps vs. Topological Maps.} DUET encodes the precise GPS coordinates of each node for graph learning. We also attempt to input these coordinates into GPT for understanding and reasoning. However, the agent appears to struggle with global  exploration based on the coordinates, resulting in a performance decline compared to the baseline that does not utilize the map. On the other hand, our proposed map-guided prompting method provides the topological relationship in a natural language form for GPT-4 and GPT-4V to understand, leading to a significant performance improvement.

\begin{table}[t] 
 \centering
 % \vspace{-2mm}
 \small
 \renewcommand\tabcolsep{2.0pt}
\resizebox{1.0\columnwidth}{!}{
\begin{tabular}{l|cc|cccc}
 \toprule
 LLMs & Map & Planning & NE$\downarrow$ & OSR$\uparrow$ & SR$\uparrow$ & SPL$\uparrow$  \\
 \midrule
 \multirow{3}{*}{GPT-4}
  & $\times$ & $\times$ & 6.49 & 49.5 & 32.9 & 19.4 \\
  & Topological & $\times$ & 6.40 & 59.7 & 37.5 & 24.8   \\
  & Topological & Adaptive & 5.80 & 61.6 & 41.2 & 25.4 \\
  \midrule
  \multirow{5}{*}{GPT-4V}
  & $\times$ & $\times$ & 5.96 & 58.8	& 42.6 & 34.7 \\
  & Coordinate & $\times$ & 6.12 & 55.1 &	41.2 & 32.8\\
  & Topological & $\times$ & 5.89	& 56.5 & 44.9 &	36.5 \\
  & Topological & Action & 5.82 &	58.3 & 45.4 & 35.6 \\
   & Topological & Adaptive & 5.62	& 57.9 & 47.7 &	38.1\\
 \bottomrule
 \end{tabular}
}
% \vspace{-2mm}
\caption{Ablation of different map or planning designs on 72 various scenes of the R2R dataset.   
}
\label{table:ablation}
 \vspace{-2mm}
\end{table}

\noindent\textbf{Global Action Planning vs. Adaptive Path Planning.}
DUET develops a global action planning, which involves selecting an optimal node from both accessible and unexplored inaccessible nodes, and teleporting the agent to that node. We also implement a similar agent that incorporates nodes from the ``Supplementary Info" into the action space for action planning by GPT-4V. Experimental results indicate that this approach does not significantly improve the zero-shot agent equipped with GPT. Instead, we adopt an adaptive planning approach, where GPT explicitly outputs a segment of multi-step path planning, allowing flexible attention to multiple potential nodes or sub-goals and the ability to correct previous errors. 
This effectively leverages the advantages of the GPT for thinking and planning, rather than selecting a single action.

\subsection{Case Study}
\label{sec:case}

In Figure~\ref{fig:case}, we showcase a successful example from REVERIE that demonstrates the various abilities of MapGPT (GPT-4V based). This example poses some challenges for a zero-shot agent since the places observed at place 4, namely places 5 and 9, as well as the previously observed places 1, 6, 7, and 8, do not contain the bathroom which is behind the door. Additionally, the instructions in REVERIE typically do not include information about turning or any other specific actions. Therefore, the agent needs to explore the entire environment to determine the correct direction.
Benefiting from map-guided prompting with adaptive planning, MapGPT demonstrates a strong understanding of the topological relationships between nodes and adaptively performs multi-step path planning. Based on six unexplored global candidates, the agent systematically conducts global exploration by selecting the four most probable nodes, as they are situated within the bedroom and are more likely to be connected to the bathroom. Besides, the planning content mentions the possibility of backtracking to place 2 for re-exploration if necessary, and is also adaptively updated upon discovering the direct connection between places 5 and 4 in the map.
After several steps, when the agent moves to place 8, it discovers the bathroom hidden behind the door and successfully reaches the destination.

\section{Conclusion}
In this paper, we propose a novel zero-shot agent, named MapGPT, for the VLN task. MapGPT utilizes map-guided prompting, which builds online constructed maps using prompts that provide GPT with node information and topological relationships to activate global exploration. Additionally, we propose an adaptive planning mechanism that enables multi-step path planning based on the map, allowing the agent to systematically explore potential objectives.  Through extensive experiments, we demonstrate that MapGPT achieves state-of-the-art zero-shot performance with global thinking and path planning capabilities.

\section*{Limitations}
Despite the significant performance gap between MapGPT and models based on pre-training and fine-tuning, zero-shot VLN still holds significant research value. 
GPT's pre-training corpus contains a large amount of real-world image data, thus demonstrating great potential in terms of generalization and sim-to-real transfer. 
However, MapGPT is only experimented within a simulator that incorporates certain ideal assumptions. Developing LLM-based agents directly in the real world and addressing various real-world challenges would be a meaningful future direction.

\section*{Acknowledgements}
This work was supported in part by CAAI-Huawei MindSpore Open Fund.

\bibliography{custom}

\appendix

\appendix
\renewcommand\thesection{\Alph{section}}

\section*{Appendices}

\section{More Details}
\label{sec:supp_details}

\subsection{Prompts}
\label{sec:supp_prompt}

\begin{figure*}[t]
\begin{center}
 \includegraphics[width=1.0\linewidth]{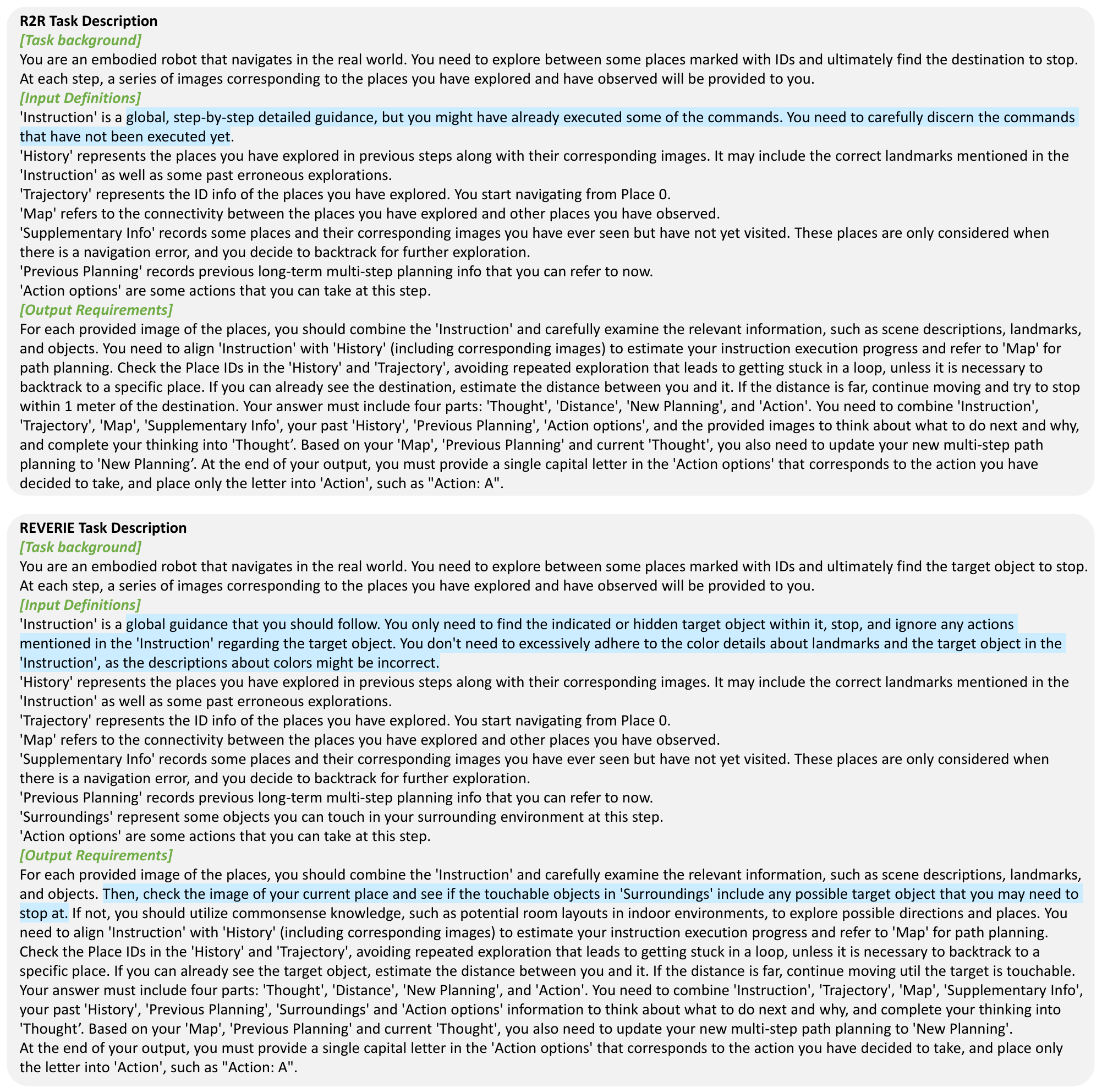}
\end{center}
% \vspace{-7mm}
  \caption{
  Task description prompts for the R2R and REVERIE datasets. We make some simple yet necessary modifications to transfer MapGPT from the R2R task to REVERIE. This work focuses on unified navigation, while instructions in REVERIE often require some interactive actions on objects. Therefore, we require the agent to ignore these actions.
  }
% \vspace{-5mm}
\label{fig:prompt}
\end{figure*}

\paragraph{Task Description}
We provide specific task description prompts that are directly fed into the system content of GPT-4V~\cite{openai2023gpt4v-system,openai2023gpt4v-technical} API. As shown in Figure~\ref{fig:prompt}, our unified prompts consist of three parts, namely task background, input definitions, and output requirements.

We have achieved effortless adaptation between fine-grained R2R~\cite{anderson2018vision} instructions and high-level REVERIE~\cite{qi2020reverie} instructions with only a few intuitive and necessary modifications. These modifications are primarily utilized for ignoring the extensive interactive actions with objects in REVERIE, since our unified agent is designed to focus on the navigation task. The stopping conditions also differ, as R2R only requires the agent to stop at the destination, while REVERIE demands the agent to check the target object in its surroundings before stopping.

\paragraph{Templates}
All of the environmental information collected at each step (including observations, history, maps, etc.) will be incorporated into the user messages of GPT.

For a two-stage prompt system, we formulate the observations that have been converted into text as $O_t=\{o_{t,i}\}_{i=1}^N$, where $N$ is the number of navigable viewpoints at step $t$. Each $o_{t,i}$  represents the observation towards a specific navigable point and is formulated using the prompt template:  

\begin{quote}
\textit{``\textless\{scene\}\textgreater,\,which\,also\,includes\,\textless\{objects\}\textgreater}''.
\end{quote}
Besides, we have defined six directional concepts, namely \textit{``go forward to''}, \textit{``turn left to''}, \textit{``turn right to''}, \textit{``turn around to''}, \textit{``go up to''},  and \textit{``go down to''}, according to the directions of navigable viewpoints. 
Thus, a complete action $a_{t}$ could be \textit{``B. turn around to Place 1: \textless a room with blue walls\textgreater, which also includes \textless bed, curtain, picture\textgreater''}.

For a one-stage system based on GPT-4V, we directly require the agent to refer to images corresponding to various places. At the beginning of the user messages, we input the image IDs and observed images in an interleaved format, such as \textit{``Image 0: <Img0> Image 1: <Img1> ...''}. 
Therefore, these images can be directly referenced in subsequent prompts. 
For example, when used in the action space, an action $a_{t}$ could be \textit{``C. turn around to Place 2 which is corresponding to Image 2''}. Prompts in history and maps also employ similar templates.

\subsection{Implementation Details}
\label{sec:supp_implementation}

We conduct experiments in the Matterport3D simulator~\cite{chang2017matterport3d}, which provides a discrete navigation environment with predefined navigable viewpoints. At each viewpoint, the agent can obtain visual observations and some connected navigable candidate viewpoints which are incorporated into prompts for GPT. Once GPT has selected one of these candidates by predicting the corresponding label in the prompts, we can convert it into the candidate's ID which can be executed in the simulator and teleport the agent to the selected viewpoint.

In this work, we have built the first GPT-4V-based~\cite{openai2023gpt4v-system,openai2023gpt4v-technical} agent in the VLN field, directly processing multimodal inputs in one stage. 
For a fair comparison with previous methods specifically designed for GPT-4~\cite{openai2023gpt4}, we have also implemented a two-stage system, where we follow NavGPT~\cite{zhou2023navgpt} and utilize BLIP-2 \cite{li2023blip} to provide a caption for the observation, and employ Faster R-CNN \cite{ren2015faster} to detect existing objects. 
Our core contributions, namely map-guided prompting and adaptive path planning, can be applied to both of these systems.
To adapt our MapGPT to the REVERIE dataset, we only make some simple yet necessary modifications, which demonstrates that our MapGPT is more unified in the VLN field.

%%%%%%%%%%%%%%%%%%%%%%%%%%%%%%%%%%%%%%%%%%%%%%%%%%%%%%%%%%%%%%%%%%

\section{More Qualitative Examples}

We provide additional successful and failure cases to qualitatively analyze the capabilities and limitations of our proposed MapGPT. 

\begin{figure*}[t]
\begin{center}
 \includegraphics[width=0.98\linewidth]{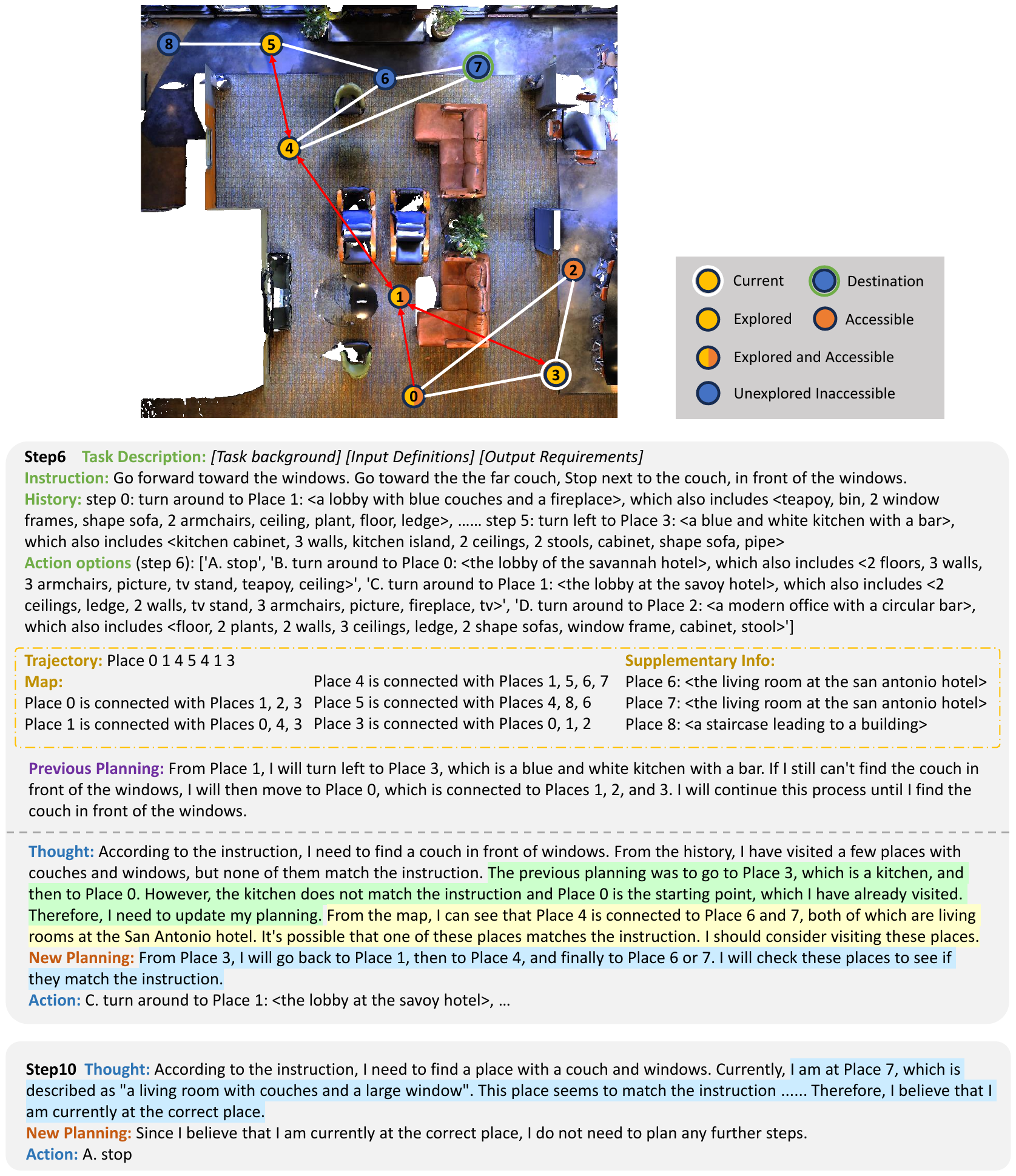}
\end{center}
\vspace{-2mm}
  \caption{
  A successful example on the R2R dataset. We demonstrate some crucial steps that leverage map-guided global exploration and planning capabilities, ultimately resulting in successful navigation.
  }
% \vspace{-5mm}
\label{fig:success1}
\end{figure*}

Figure \ref{fig:success1} demonstrates a successful case on the R2R dataset. In step 6, after thoroughly exploring places 3 and 4 connected to place 1, the agent decides to backtrack to place 1 and subsequently explore currently inaccessible places 6 and 7. Ultimately, the agent successfully terminates at place 7 in step 10.

\begin{figure*}[t]
\begin{center}
 \includegraphics[width=1.0\linewidth]{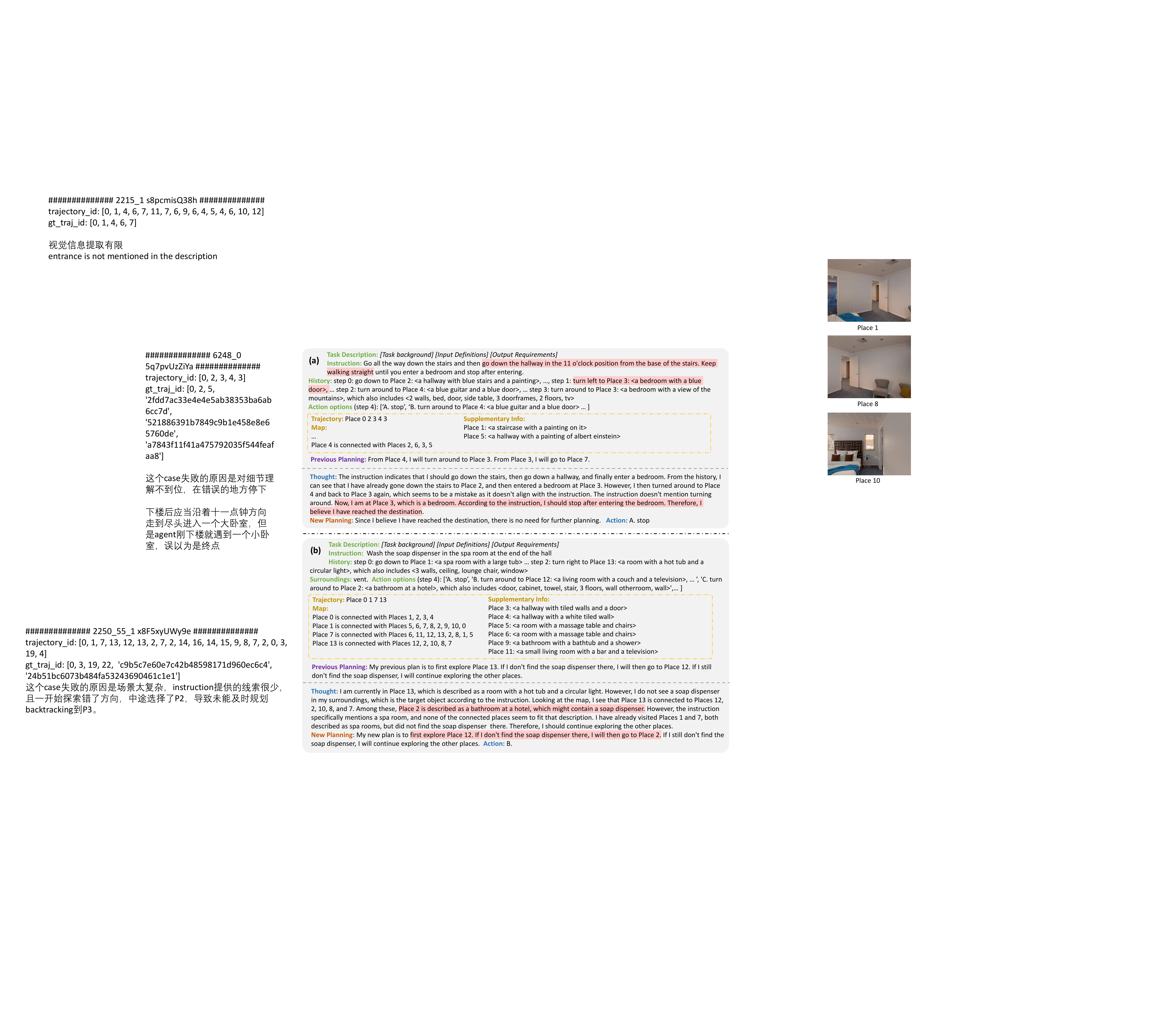}
\end{center}
% \vspace{-7mm}
  \caption{
  Two typical reasons for failure. (a) Stopping erroneously in similar locations. (b) Failure to timely backtrack and explore the correct direction if the scene is complex and the instruction does not provide much guidance (the agent has explored place 1 and plans to explore place 2, while the correct direction is 0$\rightarrow$3).
  }
\label{fig:fail}
\end{figure*}

As shown in Figure~\ref{fig:fail}, we further summarize two typical types of failure cases, which are also common challenges for other zero-shot VLN agents. (a) The agent may fail to follow the details in the instructions accurately. For instance, instead of walking straight into a bedroom in the eleven o'clock direction as instructed, it turns left in step 1 and enters another incorrect bedroom, and stops there. (b) The scenes are highly challenging, and the instructions may not provide sufficient clues. Thus, the agent may fail to explore the correct direction in time.

\end{document}